# When Post-Processing Fairness Constraints Help and When They Harm

## Evidence from Eight Cross-Domain Evaluations

**Nithin Raghava Ramachandra Narla**
Independent Researcher, Dallas-Fort Worth, Texas, United States
ORCID: 0009-0005-7458-551X
Corresponding author: nithinrf95@gmail.com

**Keywords:** algorithmic fairness; post-processing intervention; cross-domain evaluation; fairness drift; production machine learning; disparate impact

## Abstract

Fairness audits in production ML typically occur once, at deployment, on a single domain. Both fail in practice: fairness can shift after retraining or a changing user base, and interventions validated on one dataset are rarely tested across the heterogeneous domains an organization deploys. We present FAPE (Fairness Auditing for Production Environments), a four-stage framework evaluating a single post-processing intervention, Fairlearn's ThresholdOptimizer, across eight domain evaluations: criminal justice, income prediction, legal admissions, credit lending, agricultural lending, a multi-domain benchmark corpus, healthcare, and education. Each is scored on demographic parity and equalized odds difference, plus disparate impact ratio and accuracy cost where computable. Intervention effectiveness tracks baseline disparity magnitude: across model-domain pairs the constraint improved disparity in 9 of 14 high-disparity cases and worsened it in 3 of 4 near-fair ones. Each of the five high-disparity exceptions reverses under one of two measurement checks, a minimum group size or thresholds fit on held-out data. A CUSUM monitor started at deployment, tested on a simulated shift, separates constrained models that never met a 0.1 parity convention from those that met it and later regressed. A single deployment-time audit is therefore an unreliable guide, which argues for baseline-disparity screening and continuous monitoring.

---

## 1. Introduction

### 1.1 The Production Fairness Gap

Machine learning systems deployed in consequential domains, such as credit decisions, criminal justice risk assessment, agricultural lending, and educational placement, are typically audited once, at launch, and rarely monitored afterward. Sculley et al. (2015) documented how production ML systems accumulate technical debt and degrade silently over time, and Ajarra and Basu (2026) have since treated fairness auditing under model updates as a problem in its own right. AIF360 and Fairlearn, two widely used fairness toolkits, each produce a static report from a single evaluation at a single point in time. Neither asks what happens six months later, when the user base shifts, the model is retrained, or a vendor swaps the algorithm, so neither is designed to close the gap between the fairness a system had at validation and the fairness it holds in production. Consider a credit model that passes a fairness audit at launch. Six months later it is retrained, or the lender serves a different regional mix of applicants; aggregate accuracy, the metric a business dashboard tracks, holds steady while the demographic parity gap widens past the threshold that would have blocked deployment. No alert fires, because nothing measures it continuously. Stage 4 is built to detect this.

### 1.2 The Cross-Domain Generalization Problem

A second, related gap concerns generalizability. Most empirical fairness studies test an intervention on one dataset, often COMPAS, Adult Income, or German Credit, and the comparative studies that span several datasets draw on much the same small pool (Section 2.3). Whether a post-processing method that reduces demographic disparity in criminal justice risk scoring behaves the same way in agricultural lending or student performance prediction has gone largely untested. FAPE evaluates a single post-processing method across eight evaluations at once, including domains outside that pool, and finds that effectiveness varies by domain and by model in ways a single-dataset study could not surface.

### 1.3 FAPE Contributions

This paper makes four contributions. First, we present a systematic cross-domain evaluation of ThresholdOptimizer post-processing across eight deployment-context evaluations spanning criminal justice, income prediction, legal admissions, credit lending, agricultural lending, a multi-domain benchmark corpus, healthcare, and education, domains selected for their heterogeneity in scale, label distribution, and protected-attribute structure. Second, we evaluate each domain on multiple fairness and performance metrics at once, demographic parity difference and equalized odds difference throughout, plus disparate impact ratio and accuracy cost where the domain's pipeline computes them, rather than the single-metric evaluations common in prior work, which by construction cannot surface the tradeoffs between them; a domain that looks fair on one metric can look markedly unfair on another, and our design is built to catch that. Third, we apply CUSUM sequential monitoring to fairness after deployment, tested on a simulated shift, extending evaluation beyond the point-in-time audit that the open-source fairness toolkits stop at. Fourth, we report an empirical effectiveness pattern, observed across all eight evaluations, for when post-processing constraints move from beneficial to counterproductive: counting every model-domain pair, the constraint improved disparity in 9 of the 14 pairs with a baseline demographic parity difference above 0.2, or 6 of 11 to 7 of 12 once the duplicated evaluation of Section 6.4 is counted once, and worsened it in 3 of the 4 pairs below 0.05. Each of the five high-disparity exceptions reverses under one of two measurement checks, a minimum group size or thresholds fit on held-out data (Sections 6.2 and 6.4).

### 1.4 Paper Organization

Section 2 reviews related work on fairness interventions, fairness metrics, cross-domain evaluation, and production monitoring. Section 3 describes the methodology, Section 4 the experimental setup and reproducibility measures, and Section 5 the results across all eight evaluations. Section 6 discusses their implications for practitioners, and Section 7 concludes.

## 2. Related Work

### 2.1 Fairness Interventions

Algorithmic fairness interventions fall into three categories, distinguished by where in the machine learning pipeline they act. Pre-processing methods modify training data before a model ever sees it; Kamiran and Calders (2012) proposed reweighting and resampling techniques that adjust the distribution of protected attributes and outcomes to reduce discriminatory patterns in the data itself. In-processing methods instead modify the training objective directly. Zhang et al. (2018) introduced adversarial debiasing: a predictor trained jointly with an adversary that tries to recover the protected attribute from the predictor's output, with the predictor penalized each time the adversary succeeds. Both need direct access to the training pipeline, which production settings often lack: a vendor retrains on its own schedule, or an API returns predictions with no view of how the model was built. Post-processing methods avoid this problem, since they work on a model's outputs after training has already finished. Hardt et al. (2016) formalized this approach with equalized odds post-processing, applying group-specific decision thresholds to an already-trained model's scores to satisfy a fairness constraint without touching the model itself. This is the intervention FAPE evaluates, implemented via Fairlearn's ThresholdOptimizer, because it is deployable against any classifier already in production.

### 2.2 Fairness Metrics and Impossibility

A parallel body of work defines what fairness means numerically. Dwork et al. (2012) introduced individual fairness: similar individuals should receive similar predictions. Demographic parity and equalized odds (Hardt et al., 2016) take a different approach: they evaluate fairness at the group level, not the individual level. A third formulation, the disparate impact ratio, grew out of the EEOC's four-fifths rule in employment discrimination law and has since become a general convention across fairness research more broadly. Under this rule, a selection rate for a protected group below 80% of the rate for the most-favored group counts as evidence of adverse impact. Chouldechova (2017) proved that several of these metrics cannot be jointly satisfied when base rates differ across groups, except under narrow conditions. That impossibility result means intervention studies that report a single metric can obscure the tradeoff practitioners most need to see. Measurement can mislead as well: Sariola et al. (2026) found that resampling hiring data to equalize base rates appeared to reach parity on traditional measures while leaving an absolute disparity of about 10% when measured with audit-study data. Both findings motivate FAPE's evaluation design; we report demographic parity difference and equalized odds difference for every domain, alongside disparate impact ratio and accuracy cost where available.

### 2.3 Cross-Domain Fairness Evaluation

Empirical fairness research has mostly evaluated interventions one dataset at a time, most often COMPAS, Adult Income, or German Credit. Comparative studies widen the view but reuse those benchmarks. Friedler et al. (2019) compared fairness-enhancing interventions across several benchmark datasets and found them sensitive to how the data were split into training and test sets, and Chen et al. (2023) evaluated seventeen mitigation methods, equalized odds post-processing among them, on the five datasets that ship with the AIF360 toolkit: Adult, COMPAS, German Credit, Bank Marketing, and MEPS. Simson et al. (2025) addressed the narrowness of that pool with FairGround, a corpus of fairness-annotated datasets built for broader, more reproducible evaluation. FAPE applies one post-processing intervention to eight evaluations: COMPAS and MEPS from that pool, Folktables in place of Adult, and Law School (Wightman, 1998), Lending Club, SBA agricultural loan, and Student Performance (Cortez and Silva, 2008) data from outside it, with three evaluations loaded through FairGround. Section 6.4 states how many independent data sources the eight rest on.

### 2.4 Production Fairness Monitoring

A separate literature examines what happens to machine learning systems after deployment, largely independent of the fairness-metrics literature above. Sculley et al. (2015) documented that production ML systems accumulate hidden technical debt and degrade in ways that are not visible in standard accuracy monitoring. Ajarra and Basu (2026) carried the concern to fairness, deriving how many labeled samples an auditor needs to check statistical parity when a model owner may keep updating the model, and which updates leave that property intact. Tooling for checking fairness continuously has lagged. Breck et al.'s (2017) 28-test production readiness rubric asks that a model be tested for considerations of inclusion before release, but none of its seven monitoring tests covers fairness, and the open-source toolkits AIF360 and Fairlearn evaluate a model at a single point in time. Commercial monitoring does exist: Amazon SageMaker Clarify recomputes bias metrics on each scheduled window of live data and alerts when a window's confidence interval falls outside an allowed range (Amazon Web Services, n.d.). FAPE's Stage 4 is an open, sequential alternative, a CUSUM monitor that accumulates small excesses across windows rather than judging each window alone.

# 3. Methodology

## 3.1 Framework Overview

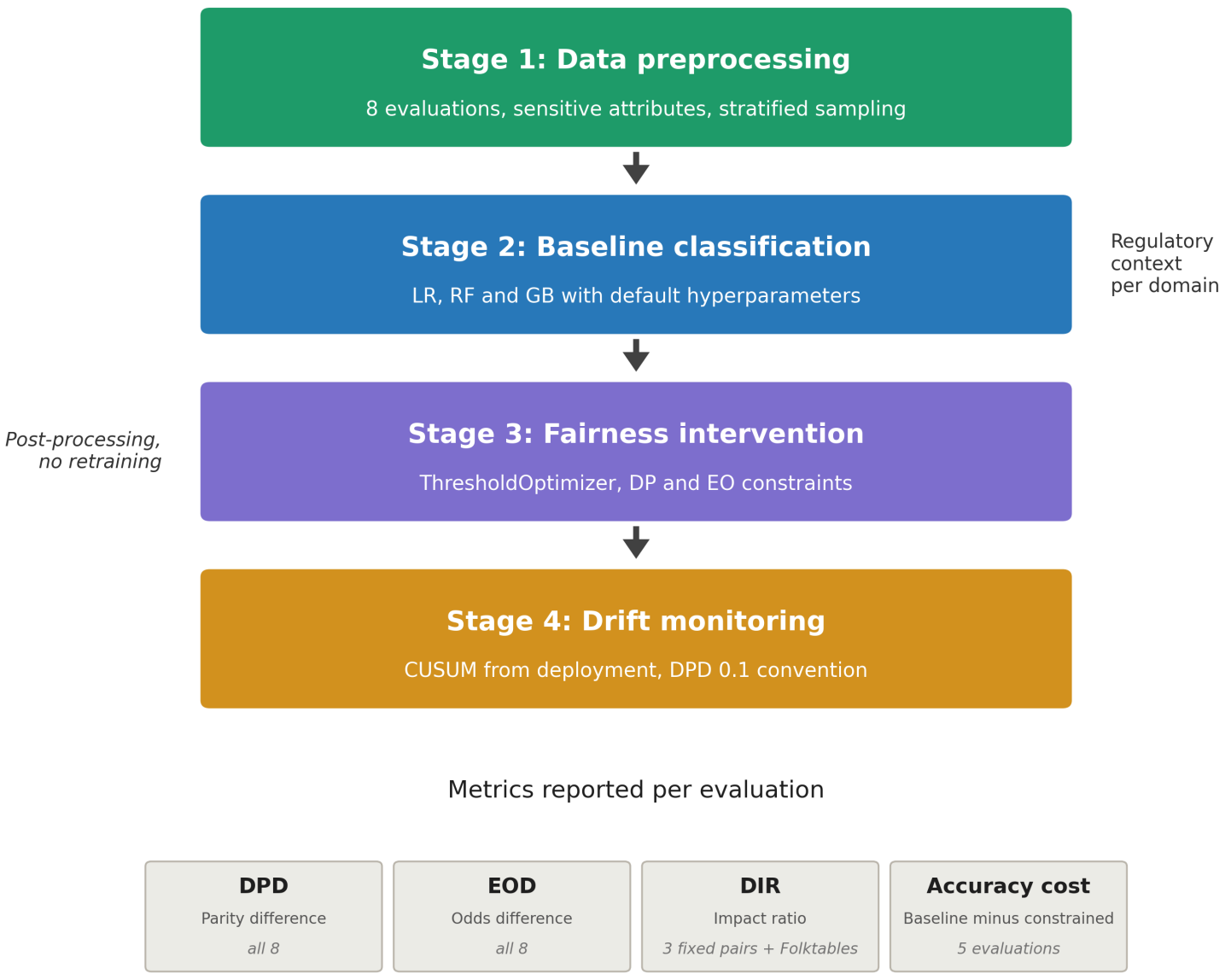

Figure 1: FAPE's four-stage pipeline

**Figure 1.** FAPE's four-stage evaluation pipeline, applied across all eight evaluations: Stage 1 (data preprocessing), Stage 2 (baseline classification), Stage 3 (fairness intervention via ThresholdOptimizer), and Stage 4 (drift monitoring).

Figure 1 shows the pipeline used throughout. FAPE runs a single post-processing fairness intervention across eight real-world evaluations in four stages: data preprocessing, baseline classification, fairness intervention, and drift monitoring. The intervention touches a model only after training, so no retraining is needed, which makes the framework usable against classifiers already in production. Every domain goes through the same four stages with the same model settings, so that differences across domains trace mainly to the domain, though the domain scripts differ in which models and metrics they carry through (Sections 3.3 and 3.5). Where a domain-specific standard applies, results are framed against it: ECOA's individual-applicant provisions for Lending Club and its business-credit provisions for Agricultural (Section 6.4). Outside the two lending domains, the 0.8 disparate impact ratio threshold is applied as a general fairness-research convention rather than a compliance test, consistent with Section 2.2. Federal civil-rights statutes may still govern deployed systems in these domains, Title VI and Title IX for federally funded education and Affordable Care Act Section 1557 for federally funded health programs among them. Whether a given deployment falls under them is a legal question this study does not settle.

### 3.2 Datasets and Domains

FAPE evaluates eight domain evaluations spanning criminal justice, income prediction, legal admissions, credit lending, agricultural lending, a cross-domain benchmark corpus, healthcare, and education, drawn from seven independent data sources for the reason given in Section 6.4. Table 1 summarizes each domain's dataset, sensitive attribute, sample size, and governing regulatory context. Domains vary along two dimensions: sensitive attribute type, with race, sex, income band, and business type each appearing at least once, and regulatory regime, since a lending model and an admissions model face different legal standards.

| Domain | Dataset | Sensitive Attribute | Sample Size | Regulatory Context |
|---|---|---|---|---|
| Criminal Justice | COMPAS (ProPublica) | Race (6 groups) | 6,172 | 0.8 DIR convention applied (Section 3.1) |
| Socioeconomic | Folktables ACS | Race | 1,589,032 (100,000 training sample) | 0.8 DIR convention applied (Section 3.1) |
| Legal Admissions | Law School Admissions | Race | 18,692 | 0.8 DIR convention applied (Section 3.1) |
| Credit Lending | Lending Club | Income band (proxy) | 2,260,701 raw / 1,348,099 filtered (100,000 training sample) | ECOA, individual-applicant provisions |
| Agricultural Lending | SBA 7(a) | Business type | 15,845 | ECOA, business-credit provisions |
| Benchmark Corpus | FairGround (law_school_lequy) | Race | 18,692 | 0.8 DIR convention applied (Section 3.1) |
| Healthcare | MEPS Panel 19 FY2015 (FairGround) | Race | 15,830 | 0.8 DIR convention applied (Section 3.1) |
| Education | Student Performance | Sex | 395 (Math), 649 (Portuguese), evaluated separately | 0.8 DIR convention applied (Section 3.1) |

**Table 1.** Summary of the eight domain evaluations, their sensitive attributes, sample sizes, and applicable regulatory context. FairGround's row gives the law_school_lequy sub-dataset used for its headline results; a second sub-dataset, creditcard, is reported under the equalized odds constraint in Section 5.3.

Two domains were sampled for computational feasibility. Folktables ACS, used here in place of the smaller and methodologically critiqued Adult Income benchmark (Ding et al., 2021), provides a full national sample across all fifty states; a 100,000-record random sample was used for model training, with the subsequent train-test split stratified on the income classification outcome. Lending Club, after filtering the original 2,260,701 records to a binary paid-off versus charged-off outcome (1,348,099 records), was further reduced to a 100,000-record stratified training sample. In both cases the models are trained and their fairness metrics computed on the sample rather than the full population. Student Performance is evaluated as two independent subjects, Math (395 records) and Portuguese (649), since the two showed materially different fairness patterns; Section 5 reports Math. Three feature choices keep label information out of the inputs. MEPS uses the 41 features FairGround documents, because its raw panel file also carries the visit counts that define the utilization label; Folktables leaves out the family income-to-poverty ratio, which contains the income being predicted; and Student Performance leaves out the two interim grades.

### 3.3 Baseline Models

Logistic regression, random forest, and gradient boosting are trained with scikit-learn's default hyperparameters at the baseline stage across all eight evaluations. Random forest is carried through the ThresholdOptimizer fairness-intervention stage in five of the eight; the three domains reporting AUC rather than accuracy (Law School, Lending Club, Agricultural) evaluate only logistic regression and gradient boosting under the intervention, a scope decision made when those three scripts were built. Where all three are compared under intervention, holding architecture and configuration constant isolates the intervention's fairness effect from model choice or tuning. No domain-specific tuning was applied. Section 5.1 compares baseline performance.

### 3.4 Fairness Intervention

The intervention evaluated is ThresholdOptimizer, Fairlearn's post-processing implementation, applied under two separate fairness constraints, demographic parity and equalized odds, and fit on each training split, choosing the thresholds that maximize balanced accuracy under each. During verification, its post-constraint outputs varied between otherwise identical runs, despite fixing random_state=42 on all three underlying classifiers. The cause turned out to be internal to ThresholdOptimizer itself: fairlearn 0.13.0's predict method performs a probabilistic threshold interpolation controlled by its own random_state parameter, left unset. Passing it explicitly to all twenty-two ThresholdOptimizer.predict calls fixed the issue, confirmed by two consecutive full runs per domain returning identical output on every metric.

### 3.5 Evaluation Metrics

FAPE reports up to four metrics per domain. Demographic parity difference and equalized odds difference are computed for all eight; disparate impact ratio and accuracy cost are computed where each domain's intervention script supports them, and Section 5 states the coverage for each. Demographic parity difference measures the absolute gap in positive prediction rate between the most and least favored groups; following common practice in the fairness literature, a difference below 0.1 is interpreted as broadly acceptable, treated as a research convention rather than a binding legal standard for any domain here. Equalized odds difference measures the corresponding gap in error rates, true positive and false positive rate, between groups, capturing disparities that demographic parity alone can miss when base rates differ. Disparate impact ratio expresses the same underlying comparison as a ratio rather than a difference. Each domain that reports one computes it as the selection rate of a designated group divided by that of a designated reference group, with the pair fixed in advance rather than chosen as the observed minimum and maximum: minority over majority for Law School, partnership over corporation for Agricultural, and lowest income quartile over highest for Lending Club. A ratio below 0.8 is applied as the convention described in Sections 2.2 and 3.1, and as an empirical heuristic throughout, including for Lending Club and Agricultural: ECOA governs lending discrimination but does not adopt the four-fifths rule, which is an EEOC employment guideline. Because the pair is fixed, a ratio far from 1.0 in either direction is disparity, and which direction is adverse depends on the outcome. Law School's outcome is bar passage, so a ratio below 0.8 means minority applicants are predicted to pass less often; both lending domains predict default, so there the adverse direction is a ratio above 1.0, and Section 5.4 reads each domain accordingly. Accuracy cost is baseline accuracy minus constrained accuracy, reported because removing disparity at a large accuracy cost is a different tradeoff from removing it nearly for free. Reporting these together, rather than only one, responds to the impossibility result and the masking effect discussed in Section 2.2.

### 3.6 Drift Detection

Stage 4 extends the evaluation beyond a single audit using CUSUM, cumulative sum, sequential monitoring, which accumulates small deviations across a sequence of observations and so catches gradual, sustained drift that a single comparison would miss. For each model, 30 simulated observations cover three versions: the baseline, the constrained model that is deployed, and a synthetic shift that moves demographic parity difference linearly from its post-constraint value 60% of the way back toward baseline. The size of each simulated regression is therefore fixed by construction as a share of that model's correction. Monitoring starts at deployment, since the baseline model never goes live: the score accumulates any excess of demographic parity difference over the 0.1 convention of Section 3.5 plus a 0.01 slack, and an alert fires once that accumulated excess passes 0.1. This targets the failure mode identified in Section 2.4 as a proof of concept; no production drift was observed.

## 4. Experimental Setup

### 4.1 Implementation

All experiments were implemented in Python 3.11.9, using scikit-learn 1.9.1 for baseline model training and Fairlearn 0.13.0 for the ThresholdOptimizer intervention. The random forest and gradient boosting models of Section 3.3 are themselves ensemble learners, bagging and boosting; Section 6.4 returns to the separate question of constraining a combined ensemble. Random seed 42 was set globally and passed to every ThresholdOptimizer.predict call, for the reason given in Section 3.4. The sampling applied to Folktables ACS and Lending Club, and the separate treatment of Student Performance's two subjects, are described in Section 3.2. The Law School Tensorflow endpoint returned a 403 error, so both the standalone Law School domain and the FairGround row load law_school_lequy from the corpus instead, as provided and without modification. Section 6.4 gives the consequence for the domain count.

### 4.2 Reproducibility

All code, data loaders, model outputs and figures are at github.com/nithinnarla/fape-fairness-ml. Every dataset used is publicly available: COMPAS recidivism data via ProPublica (Angwin et al., 2016), Folktables ACS via the folktables Python package, Law School, FairGround and MEPS data via the FairGround corpus, Lending Club loan data via Kaggle, SBA 7(a) agricultural loan data via the U.S. Small Business Administration public data portal, and Student Performance via the UCI Machine Learning Repository. The repository contains 83 EDA, 68 baseline and 79 intervention and monitoring figures, committed with the code that generated them, so every reported result traces to executable code. Exact versions are pinned in requirements.txt, and all seven domain scripts were re-run from scratch in a fresh virtual environment built from that file. Every metric value in Section 5 reproduced exactly, including Table 2, the baseline accuracy and AUC values, the disparate impact ratios and group rates of Section 5.4, and the two checks of Sections 6.2 and 6.4.

## 5. Results

### 5.1 Baseline Model Performance

No single model dominates baseline performance. Among the five evaluations where true classification accuracy is available, gradient boosting achieves the highest accuracy in three (Folktables, 0.756; Student, 0.658; MEPS, 0.859) while logistic regression leads in the other two (COMPAS, 0.686 vs. gradient boosting's 0.674; FairGround's law_school_lequy sub-dataset, 0.913 vs. 0.910). Student shows the lowest accuracy of these five, followed closely by COMPAS; Student is the smallest dataset in the study. Random forest is never the top performer in any domain, in either the accuracy-reporting or AUC-reporting group.

The remaining three domains report AUC rather than accuracy, since their intervention scripts do not compute an accuracy score for the baseline model. Gradient boosting leads all three here as well: Law School (0.878) and Lending Club (0.712) both show only a small gap over logistic regression (0.872 and 0.706 respectively), while Agricultural shows by far the largest gap of the group (0.938 for gradient boosting, 0.727 for logistic regression) alongside the highest AUC observed across the full study. Accuracy and AUC are not comparable, so the two groups are ranked separately.

Table 2 gives the full metric grid underlying Sections 5.2 and 5.3.

| Evaluation | DPD LR | DPD RF | DPD GB | EOD LR | EOD RF | EOD GB |
|---|---|---|---|---|---|---|
| COMPAS | 0.545→0.714 | 0.568→0.714 | 0.857→0.571 | 0.701→0.654 | 0.686→0.731 | 1.000→0.659 |
| Folktables | 0.301→0.348 | 0.290→0.193 | 0.302→0.373 | 0.728→0.314 | 0.732→0.836 | 0.767→0.417 |
| Law School | 0.329→0.011 | n/e | 0.351→0.030 | 0.543→0.060 | n/e | 0.528→0.007 |
| Lending Club | 0.018→0.019 | n/e | 0.024→0.018 | 0.038→0.047 | n/e | 0.053→0.049 |
| Agricultural | 0.005→0.016 | n/e | 0.009→0.031 | 0.005→0.047 | n/e | 0.073→0.177 |
| FairGround | 0.329→0.010 | 0.336→0.012 | 0.342→0.014 | 0.543→0.061 | 0.524→0.472 | 0.518→0.016 |
| MEPS | 0.069→0.013 | 0.089→0.253 | 0.092→0.013 | 0.034→0.030 | 0.053→0.071 | 0.056→0.039 |
| Student (Math) | 0.212→0.010 | 0.235→0.363 | 0.237→0.215 | 0.204→0.188 | 0.263→0.180 | 0.314→0.114 |

**Table 2.** Demographic parity and equalized odds difference, baseline to post-constraint, for every model-domain pair. n/e marks a model not evaluated under the intervention in that domain (Section 3.3). Generated by src/make_results_table.py from the source the figures use.

### 5.2 Post-DP Constraint: DPD Results

Applying the demographic parity constraint produces model-dependent outcomes. FairGround's law_school_lequy sub-dataset shows the strongest result, all three models cutting DPD by 96 to 97%, and Law School's two models, on the same underlying data, reach 96.7% and 91.5%.

In three domains the constraint works against the intended goal for most or all evaluated models. Agricultural worsens under both models evaluated. COMPAS improves only under gradient boosting, its other two models worsening to an identical 0.714, and Folktables only under random forest; Section 6.2 shows that both domains' values are set by groups of fewer than 30 test records.

MEPS, the healthcare domain, is the only evaluation whose baseline disparity falls between the two thresholds this study set. Logistic regression and gradient boosting improve by 81% and 86%, while random forest's DPD nearly triples, from 0.089 to 0.253, a worsening Section 6.4 traces to the data its thresholds were fit on.

Lending Club, near-fair at baseline, splits by model: gradient boosting improves modestly while logistic regression worsens slightly. Student splits too, logistic regression and gradient boosting improving while random forest worsens by more than half. Models in one domain responding in opposite directions to the same constraint recurs across the study and motivates the domain-and-model-specific framing in Section 6.

### 5.3 Post-EO Constraint: EOD Results

Under the equalized odds constraint, Law School shows the strongest result, both models improving and gradient boosting by 98.7%. All three of Student's models improve, gradient boosting by 63.7%.

FairGround's two evaluated sub-datasets diverge sharply. On law_school_lequy, used for FairGround's headline comparisons elsewhere, logistic regression and gradient boosting improve by 89% and 97% while random forest improves by 10%. The creditcard sub-dataset shows the opposite, all three worsening from baselines between 0.011 and 0.018.

The remaining domains are mixed. COMPAS, Folktables and MEPS each improve under logistic regression and gradient boosting and worsen under random forest: COMPAS's gradient boosting falls from 1.000 to a still high 0.659, Folktables improves by 57% and 46%, and MEPS by 12% and 30%, while its random forest rises from 0.053 to 0.071. Agricultural worsens under both models. Lending Club barely moves in either direction.

### 5.4 Disparate Impact Ratio

Disparate impact ratio is computed per domain using each domain's own sensitive-attribute groups and is not aggregated, since the compared groups differ by domain. Law School's gradient boosting moves DIR from 0.643 to 0.957, crossing the 0.8 convention of Section 3.5 after intervention (Figure 2).

Both lending domains predict default, so a ratio above 1.0 means the designated group is flagged more often. Lending Club's lowest income quartile is flagged 2.778 times as often as the highest under gradient boosting and 1.832 times under logistic regression, amplifying a gap of 1.44 times in the observed default rates. The constraint brings those ratios to 0.973 and 0.992, but not by flagging low-income borrowers less: gradient boosting's predicted default rate rises from under 4% to about 40% in every quartile, twice the 20% default rate in the data. Agricultural shows the same mechanism at a smaller scale. Gradient boosting's partnership-to-corporation ratio moves from 0.653 to 1.042, just past parity, as every business type's predicted default rate rises from about 2% to between 16 and 19%, against a 5% default rate.

Folktables reports DIR by race against White under the equalized odds constraint, for each group of at least 30 test records, rather than for one fixed pair. Under gradient boosting five groups start below 0.8 and four cross it: Black respondents (0.72 to 0.83), multiracial respondents (0.70 to 0.92), American Indian respondents, a group of 214 (0.61 to 1.00), and Pacific Islanders, a group of 33 (0.72 to 1.03). The lowest, Other, rises from 0.42 to just under 0.8. COMPAS, FairGround and MEPS compute no fixed-pair DIR, and Student computes one only at baseline, female over male, 0.49 in Math; a missing ratio is distinct from a below-threshold result. A ratio can reach parity while a model's treatment of every group changes sharply, so DIR is read here alongside the rates behind it.

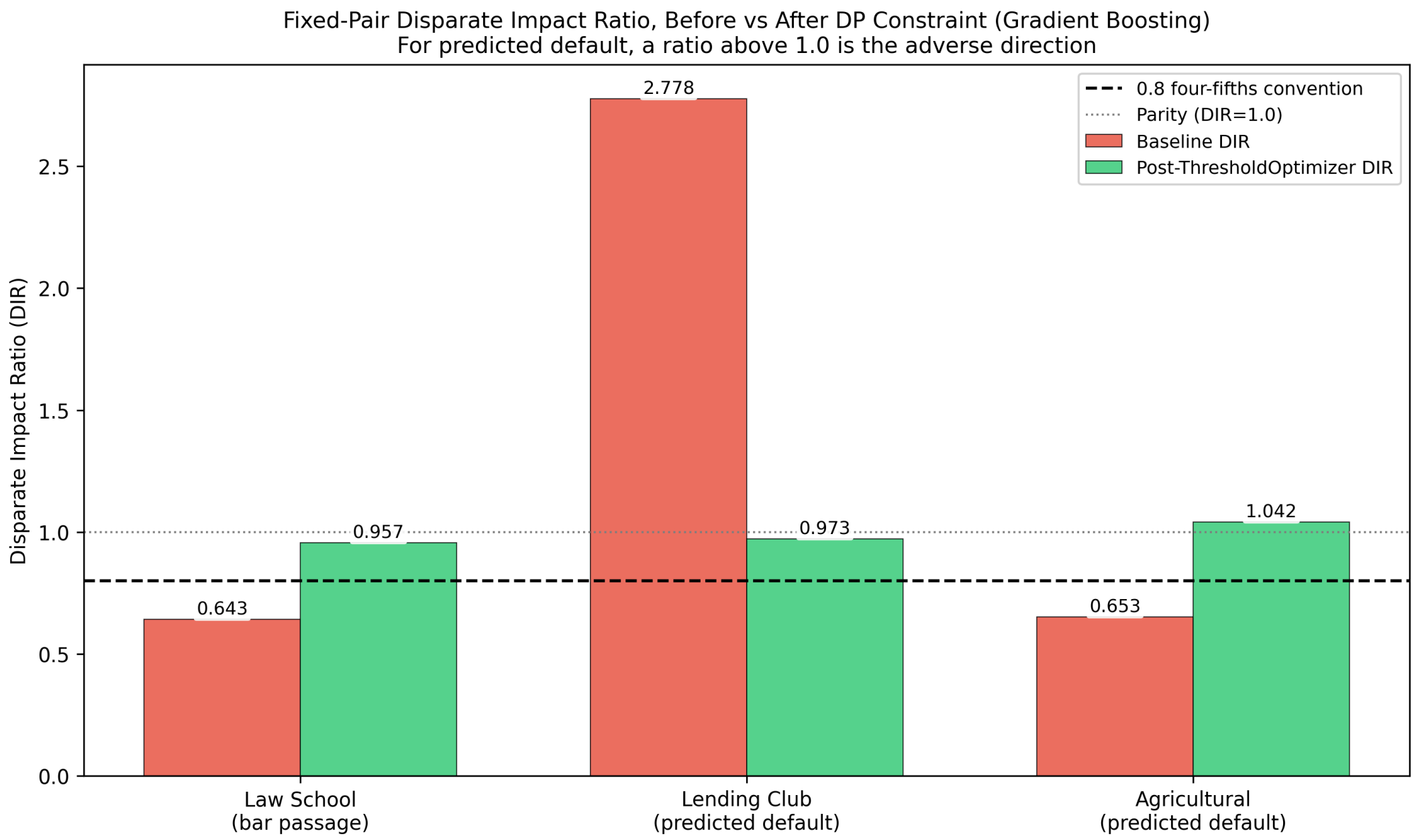

Figure 2: DIR before and after ThresholdOptimizer, three domains

**Figure 2.** Disparate impact ratio before and after ThresholdOptimizer for the three domains with a fixed-pair ratio at both points. Law School moves from below to above the 0.8 convention line. Both lending domains predict default, and their ratios approach parity because every group's predicted default rate rises (Section 5.4).

## 5.5 Cross-Domain Comparison

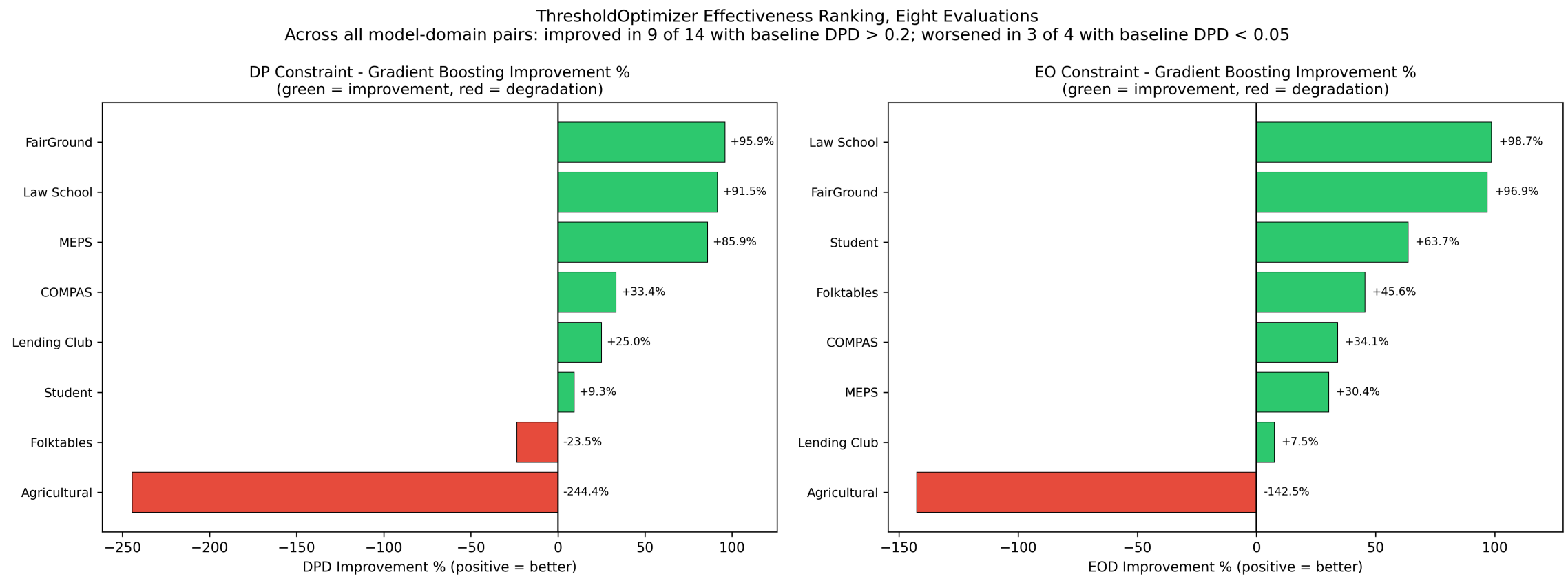

Figure 3: DP and EO improvement ranking across eight evaluations

**Figure 3.** Gradient boosting's DP and EO improvement percentage across the eight evaluations, ranked low to high. Agricultural worsens under both constraints and Folktables under demographic parity; the other six improve under both.

Considered together, the results show a pattern, with exceptions, in when ThresholdOptimizer helps (Figure 3). Counting every model-domain pair the study evaluated, rather than one representative model per domain, the constraint improved DPD in 9 of the 14 pairs whose baseline DPD exceeded 0.2, and worsened DPD in 3 of the 4 pairs whose baseline fell below 0.05. Five of those 14 use law_school_lequy, evaluated twice (Section 6.4): three via FairGround, two via Law School. Counting it once gives 6 of 11 or 7 of 12 depending on which pipeline is retained; the near-fair count is unaffected. Four of the five high-disparity exceptions, in COMPAS and Folktables, disappear when groups of fewer than 30 test records are set aside (Section 6.2), and the fifth, Student's random forest, when thresholds are fit on held-out data (Section 6.4). Read at the domain level using gradient boosting, the same pattern holds with different arithmetic: effective in four of five high-baseline domains, with Folktables the exception, and counterproductive in one of two near-fair domains, with Lending Club the exception. The pair-level count is the more complete statement, since it uses every measurement.

MEPS occupies the range between the two thresholds, baseline DPD 0.069 to 0.092, which no other evaluation reaches. Two of its three model-domain pairs improved; the third is random forest, whose worsening reverses with held-out thresholds. That is the only evidence this study offers about the middle of the range, and three pairs from a single data source cannot establish a third threshold; it is reported as an observation that the untested middle behaved more like the high-disparity group than the near-fair one, and as a target for future work.

Effectiveness also varies by model within a domain, though the checks in Section 6 show that much of this variation is measurement rather than model behavior: COMPAS's split between gradient boosting and its other two models disappears on groups of at least 30 records, and random forest's worsening in Student and MEPS reverses with held-out thresholds. No single model dominates across every domain and constraint. Gradient boosting is not uniformly the strongest baseline performer, and it does not reliably deliver the largest fairness gain either: in FairGround it improves least of the three while paying the largest accuracy cost (0.159).

### 5.6 Accuracy-Fairness Tradeoff

Accuracy cost is not computable for Law School, Lending Club, and Agricultural, whose intervention scripts report AUC rather than accuracy, a scope limitation of those scripts. Among those three, Law School shows by far the largest DPD improvement, 96.7% and 91.5%.

Among the five evaluations where accuracy cost is computable, FairGround shows the clearest tradeoff: its three models cut DPD by nearly the same 96 to 97%, yet gradient boosting pays 0.159 in accuracy and logistic regression 0.148, while random forest pays 0.010, a counterexample to the idea that a large fairness gain must be paid for. MEPS pays between 0.071 and 0.133. Student's gradient boosting pays 0.076 for a modest improvement, and its random forest pays the same while DPD worsens. COMPAS and Folktables cost 0.035 or less, and COMPAS's gradient boosting is the one case where constrained accuracy slightly exceeds baseline. No single model is cheapest across every domain. These costs also include a change of decision objective, set out in Section 6.4.

### 5.7 Drift Detection Results

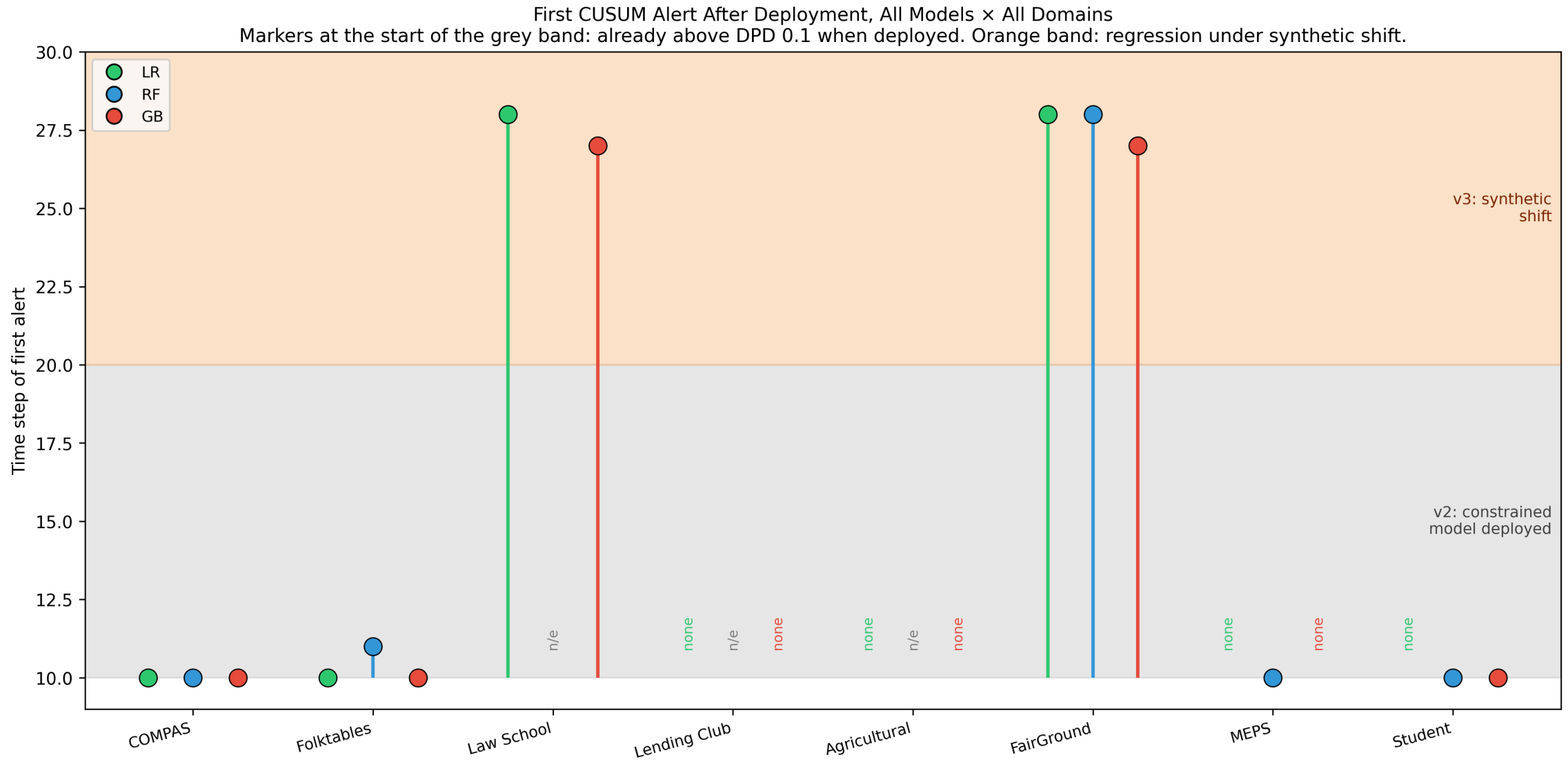

Figure 4: First CUSUM alert after deployment, by domain and model

**Figure 4.** First CUSUM alert after deployment, by domain and model. Markers at the start of monitoring are constrained models already above the 0.1 convention when deployed; markers in the shift window are regressions detected under the simulated shift; n/e marks a model not evaluated.

Monitoring from deployment sorts the 21 evaluated model-domain pairs into three groups (Figure 4). Nine are flagged within two steps of deployment because the constraint never brought them under 0.1: all three COMPAS and Folktables models, MEPS's random forest, and Student's random forest and gradient boosting. Five met the convention after the constraint and are flagged seven to eight steps into the shift: Law School's two models and FairGround's three, which Section 6.4 shows draw on the same law_school_lequy data. The other seven are never flagged: the Lending Club and Agricultural models, MEPS's logistic regression and gradient boosting, and Student's logistic regression, whose simulated regression from 0.010 peaks near 0.13 without accumulating enough excess inside the window. Because each regression is sized by construction as a share of its model's correction (Section 3.6), these groups describe how the monitor behaves, not which domains are more fragile in practice.

## 6. Discussion

### 6.1 The Effectiveness Pattern

Across the eight evaluations, a domain's baseline disparity is a strong but imperfect guide to ThresholdOptimizer's effect on that disparity. Domains with substantial baseline unfairness generally show large DPD and EOD reductions after the constraint is applied; Law School and FairGround are the clearest examples, every evaluated model improving on both metrics. Folktables is the exception at the domain level: gradient boosting's DPD rose from 0.302 to 0.373, yet on the groups with at least 30 test records it fell to 0.078 (Section 6.2). Agricultural sits at the opposite end and is the clearest case of the other half of the pattern. Its baseline DPD is 0.009, near-parity by any conventional threshold, yet the constraint raises it to 0.031, measurably worse than doing nothing. Lending Club, also near-fair, is the exception at this end, improving modestly under gradient boosting. These exceptions matter because a constraint designed to reduce disparity can increase it where little existed, and the baseline signal that anticipates this in most evaluations does not in all. The pattern is plotted in Figure 5.

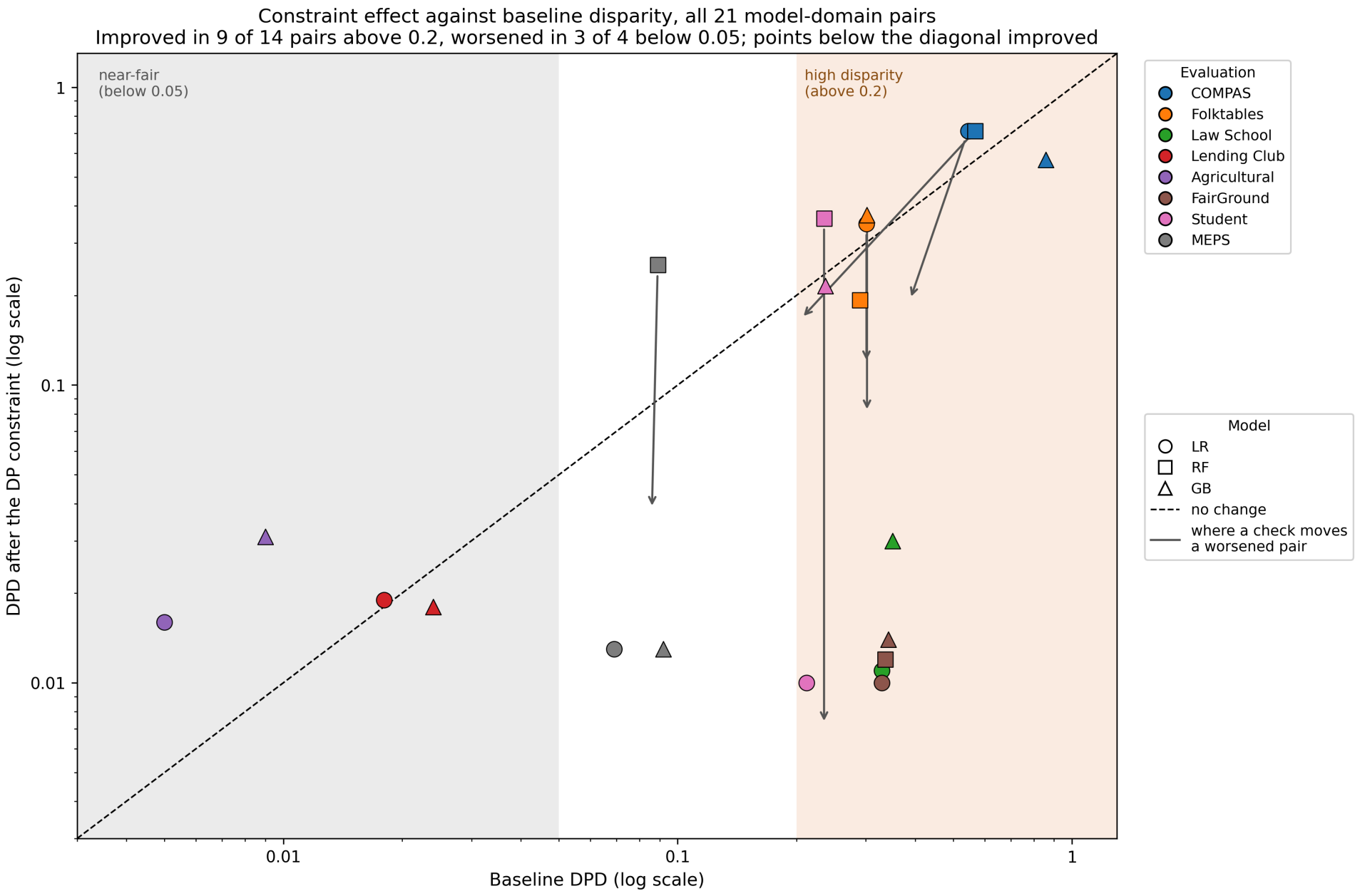

Figure 5: Baseline disparity against disparity after the constraint

**Figure 5.** Baseline DPD against DPD after the demographic parity constraint, all 21 evaluated pairs, log scales. Points below the diagonal improved; bands mark the pairs above 0.2 and below 0.05 the heuristic counts, and each arrow carries a worsened pair to where a check places it.

The practical implication is that baseline DPD should be audited before a post-processing constraint is applied. A practitioner applying ThresholdOptimizer uniformly across a portfolio, without checking whether each model's baseline disparity warrants it, risks Agricultural's outcome: introducing the exact harm the constraint exists to prevent. We propose a decision heuristic as a starting point: baseline DPD above 0.2 predicted effective intervention in 9 of 14 model-domain pairs, or 6 of 11 to 7 of 12 once the duplicated evaluation is counted once (Section 5.5), and baseline DPD below 0.05 predicted counterproductive intervention in 3 of 4. The high-disparity exceptions do not survive two checks, since 13 of 14 pairs improve on groups of at least 30 test records and the last exception reverses with held-out thresholds, while the near-fair exception, Lending Club's gradient boosting, remains. The heuristic should therefore guide where to look rather than replace checking each model's post-constraint result, on adequately sized groups, before deployment.

## 6.2 Multi-Metric Tradeoffs

When base rates differ across groups, fairness criteria conflict. Chouldechova (2017) showed that calibration and equal error rates cannot generally hold together, and demographic parity and equalized odds likewise cannot both hold for a classifier that carries information about the outcome. The results show how far the two can diverge in practice. Student's gradient boosting cuts EOD by 64% under the equalized odds constraint but DPD by only 9% under demographic parity, and MEPS's logistic regression does the reverse, 81% on DPD against 12% on EOD. A paper reporting only one metric would present each as either a clear success or a near-null result.

A single ratio can mislead in a different way. Lending Club's gradient boosting DIR falls from 2.778 to 0.973, which read alone looks like the constraint removing a near-threefold disparity against low-income borrowers. The quartile rates behind it show parity reached by predicting default for about 40% of borrowers in every quartile, low and high alike, and Agricultural repeats the pattern with a DIR within 5% of parity while its demographic parity difference triples. A paper reporting DIR alone would show two clean successes.

Small groups raise a measurement problem rather than a theoretical one. Both difference metrics take the gap between the most and least favored groups with no minimum group size. COMPAS's test set holds seven Asian defendants and one Native American defendant, and Folktables' holds five Alaska Native respondents and 25 in a combined American Indian and Alaska Native category. Every COMPAS value in Table 2, and every Folktables value except two baselines, has one of those groups at an extreme; COMPAS gradient boosting's baseline equalized odds difference of 1.000 is 0.296 without them. On the groups with at least 30 test records, which hold 1,227 of COMPAS's 1,235 test records and 19,970 of Folktables' 20,000, the demographic parity constraint narrows the gap for all six models. For logistic regression, random forest and gradient boosting the values are 0.385 to 0.187, 0.202 to 0.163 and 0.361 to 0.200 in COMPAS, and 0.301 to 0.114, 0.276 to 0.177 and 0.302 to 0.078 in Folktables. Measured that way, the high-disparity tally in Section 5.5 rises from 9 of 14 to 13 of 14, or to 10 of 11 and 11 of 12 counting law_school_lequy once. src/group_size_check.py reproduces these figures.

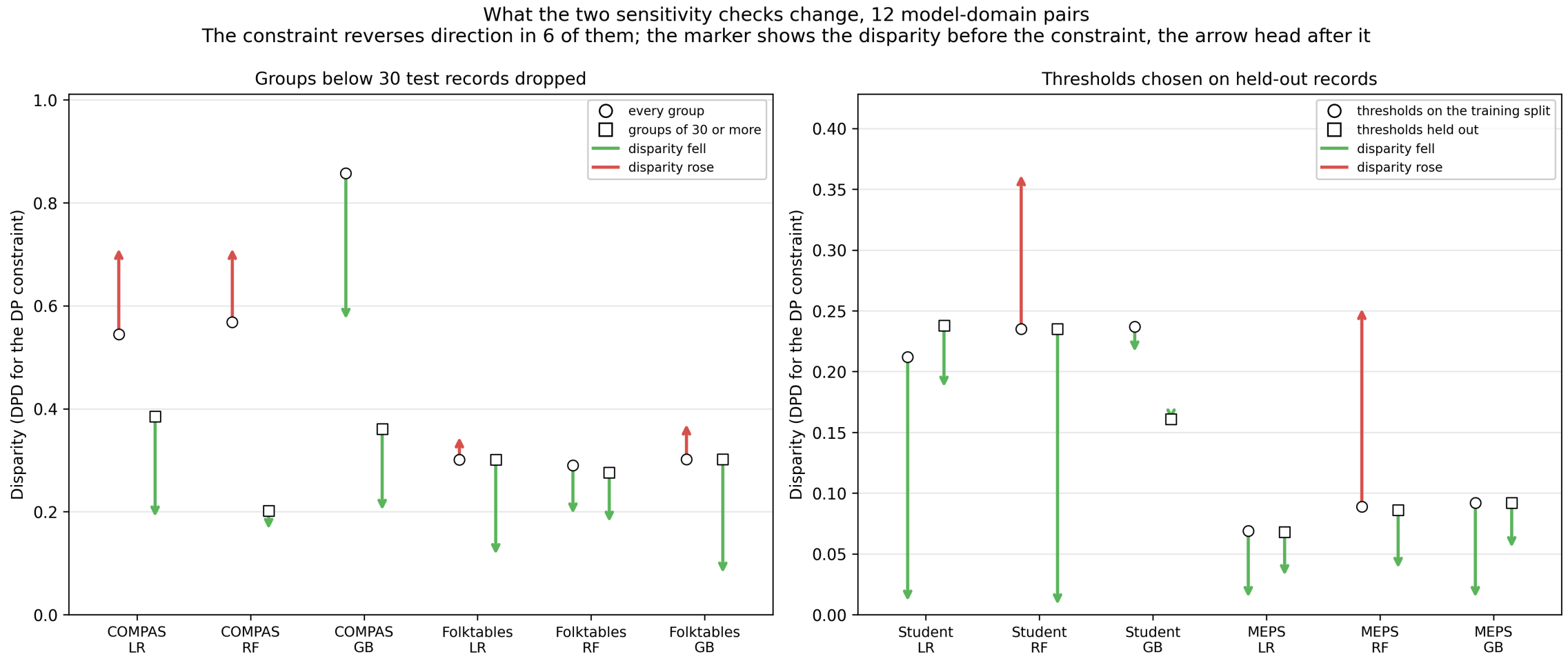

Figure 6: Disparity under each check

**Figure 6.** Demographic parity difference before the constraint (marker) and after it (arrow head), under the design the intervention scripts use and under each check. Left, groups below 30 test records dropped; right, thresholds chosen on held-out records. The constraint reverses direction in six of the twelve pairs.

### 6.3 Production Monitoring Implications

The fairness improvements in Section 5 describe a single point in time, the moment each constrained model was evaluated. Section 5.7 shows two ways that moment can mislead once a model is monitored. Nine constrained models never met the 0.1 convention at all, and a monitor running from deployment flags them within two steps, including models an audit would record as improved, such as Folktables' random forest at 0.193. Five others met the convention and then regressed under the simulated shift, whose size is fixed by construction (Section 3.6), and the monitor flagged each within eight steps of the shift beginning. We therefore recommend that CUSUM-based monitoring, or a comparable alternative, accompany any post-processing fairness intervention in production. Validating that recommendation against real deployment data is necessary future work.

### 6.4 Limitations

Agricultural lending, evaluated via SBA 7(a) loans with business type as the sensitive attribute, differs in kind from the other domains. COMPAS, Folktables, Law School, FairGround, MEPS, and Student concern individuals grouped by race or sex, and Lending Club, though it uses income band, remains consumer credit under ECOA's individual-applicant provisions. Agricultural alone concerns business entities, under ECOA's business-credit provisions, and business type is neither an immutable personal characteristic nor a proxy for one in the way income band can be. The domain also carries weight for the central claim: its baseline DPD, 0.005 and 0.009, is the lowest in the study, the constraint raises gradient boosting's to 0.031, and it supplies two of the three near-fair pairs in which the constraint backfires.

The before-and-after comparisons change two things at once. Baseline models use scikit-learn's default decision threshold, while ThresholdOptimizer chooses thresholds that maximize balanced accuracy under each constraint, so the accuracy costs in Section 5.6 and rate shifts such as Lending Club's in Section 5.4 include that change of objective and the constraint. Separating the two would need an unconstrained balanced-accuracy baseline, which this study did not run. Nor was a minimum group size applied to the reported metrics, the limitation Section 6.2 quantifies for COMPAS and Folktables.

The thresholds were also chosen on each training split, the records the base model had already been fit to, as in Fairlearn's documented examples. That suits logistic regression and gradient boosting better than random forest, which fits its training records almost perfectly (training accuracy 1.000 in Student and 0.996 in MEPS), so thresholds chosen on them transfer poorly to new records. Chosen instead on a held-out quarter of the training split, random forest's DPD under the demographic parity constraint falls from 0.235 to 0.007 in Student and from 0.086 to 0.037 in MEPS, where the reported runs worsen it to 0.363 and 0.253 (Figure 6). For logistic regression and gradient boosting the held-out design gives smaller improvements in all four cases, so it is a different estimate rather than a uniformly better one; src/threshold_holdout_check.py reproduces these figures.

Healthcare enters this study only through MEPS Panel 19, a survey-based dataset with race as the sensitive attribute, rather than through clinical records. MIMIC-III was chosen during design as the clinical dataset for the domain where Obermeyer et al. (2019) documented one of the most consequential fairness failures in a deployed algorithm, but its PhysioNet credentialed access was not granted before writing began, an access limitation rather than a design decision. Extending FAPE to clinical records once access is available remains a direct next step.

FAPE evaluates the constraint against each classifier separately rather than against a combined ensemble. Whether a post-processing constraint behaves differently on an ensemble than on its components is a question this design cannot answer and a direct extension of it.

Two of the eight evaluations draw on the same data. The standalone Law School domain and FairGround's reported sub-dataset are both law_school_lequy from the FairGround corpus, selected by that identifier in lawschool_loader.py and in the FairGround pipeline respectively, which is why both report 18,692 records. FAPE therefore covers seven independent data sources across eight evaluations, and the domain count should be read that way.

What separates the two evaluations is preprocessing, since the data are identical. Both constrain on the same binary race attribute, but the standalone pipeline keeps race among the model's inputs while the FairGround pipeline removes it, and the FairGround pipeline evaluates three models where the standalone pipeline evaluates two. The resulting numbers diverge modestly: baseline DPD under gradient boosting is 0.351 in the standalone pipeline and 0.342 in the FairGround pipeline, with post-constraint values of 0.030 and 0.014. Read as a pair, they give a small unplanned measurement of how much preprocessing alone moves these metrics on identical data, which is not a cross-domain result and is not counted as one.

## 7. Conclusion

One-time, single-domain fairness audits are the default practice in production ML, and this paper's results suggest both defaults are unsafe. Evaluating ThresholdOptimizer across eight evaluations from seven independent sources surfaces a pattern no single-domain study could show: the intervention's effectiveness tracks a domain's baseline disparity, and the high-disparity exceptions trace to small groups or to where thresholds were fit. Agricultural lending's near-fair baseline (DPD 0.009) is the clearest demonstration that applying a fairness constraint without first checking whether it is warranted can introduce the exact harm the constraint exists to prevent.

The drift results point the same way: monitoring from deployment flags constrained models that never met the 0.1 convention and, under a simulated shift, catches models that met it and then regressed. An audit performed once at launch cannot vouch for the model six months later, just as an audit on one domain cannot vouch for another.

FAPE does not resolve the impossibility results Chouldechova and others have established; no post-processing method can. What it offers instead is a practical decision framework: audit baseline disparity before intervening, report several metrics on groups large enough to measure, and monitor continuously rather than once. Several limitations bound these claims, chiefly that the drift results are proof-of-concept on synthetic shift rather than observed production data, that each before-and-after comparison changes the decision objective and adds the constraint, and that the healthcare evaluation rests on a single MEPS panel with race as its only sensitive attribute. Validating the drift-monitoring recommendation against real deployment data, and testing whether a post-processing constraint behaves differently on a combined ensemble than on its components, are the most direct next steps this work identifies.

## Declarations

**Competing interests.** None declared.

**Funding.** This research received no external funding.

**Ethics.** The study used public secondary datasets with no direct personal identifiers and no human participants, so no review board approval was required.

**Data availability.** Code, model outputs and figures: github.com/nithinnarla/fape-fairness-ml. Datasets are public and listed in Section 4.2; no new data were generated.

**Generative AI.** The study design, the four-stage framework and the choice of all eight evaluations are the author's own; every reported result comes from the code in the repository above. During manuscript preparation the author used Claude (Anthropic) to edit and condense prose, reorganize how Sections 5 and 6 present results, audit reported values against the pipeline's output and the cited sources, and assist with code: the scripts behind the figures and Table 2, the Stage 4 revision that starts monitoring at deployment, the MEPS and Folktables feature selection of Section 3.2, and the two checks of Sections 6.2 and 6.4. The author reviewed and edited all output, verified every reported value, and takes full responsibility for this publication.